%% file: main.tex
\documentclass[conference]{IEEEtran}

\usepackage{cite}
\usepackage{amsmath, amssymb, amsfonts}
\usepackage{graphicx}
\usepackage{textcomp}
\usepackage{xcolor}

\def\BibTeX{{\rm B\kern-.05em{\sc i\kern-.025em b}\kern-.08em
    T\kern-.1667em\lower.7ex\hbox{E}\kern-.125emX}}

\usepackage[linesnumbered,ruled,vlined]{algorithm2e}
\usepackage{subfig}
\usepackage{color,soul}
\usepackage{array}
\usepackage{booktabs}
\usepackage{footnote}
\makesavenoteenv{tabular}
\makesavenoteenv{table}
\usepackage[linesnumbered,ruled,vlined]{algorithm2e}
\usepackage[noend]{algpseudocode}

\usepackage[utf8x]{inputenc}
\usepackage{textcomp}

\usepackage{mathtools}
\DeclarePairedDelimiter\ceil{\lceil}{\rceil}

\begin{document}
\title{CSM-NN:  Current Source Model Based Logic Circuit Simulation - A Neural Network Approach}

\input{authors.tex}

\maketitle
\thispagestyle{plain}
\pagestyle{plain}

\input{abstract.tex}


\IEEEpeerreviewmaketitle

\input{introduction.tex}
\input{csm.tex}
\input{nn_modeling.tex}
\input{experiment2.tex}
\input{conclusion.tex}
\vspace{-2mm} 
\section*{Acknowledgement}
This research was sponsored in part by a grant from the Software and Hardware Foundations (SHF) program of the National Science Foundation. The authors would also like to thank Soheil Nazar Shahsavani and Mahdi Nazemi (of the University of Southern California) for helpful discussions.
\vspace{-2mm} 
\bibliographystyle{IEEEtran}
\bibliography{IEEEabrv,references}

\end{document}

%% file: authors.tex
\author{
\IEEEauthorblockN{Mohammad Saeed Abrishami, Massoud Pedram, and Shahin Nazarian\\}
\IEEEauthorblockA{Ming Hsieh Department of Electrical and Computer Engineering\\
Viterbi School of Engineering, University of Southern California \\ 
Los Angeles, CA 90089 \\
\{abri442, pedram, shahin.nazarian\}@usc.edu}}


%% file: abstract.tex
\begin{abstract}



The miniaturization of transistors down to 5nm and beyond, plus the increasing complexity of integrated circuits, significantly aggravate short channel effects, and demand analysis and optimization of more design corners and modes. 
Simulators need to model output variables related to circuit timing, power, noise, etc., which exhibit nonlinear behavior. 
The existing simulation and sign-off tools, based on a combination of closed-form expressions and lookup tables are either inaccurate or slow, when dealing with circuits with more than billions of transistors. 
In this work, we present CSM-NN, a scalable simulation framework with optimized neural network structures and processing algorithms. CSM-NN is aimed at optimizing the simulation time by accounting for the latency of the required memory query and computation, given the underlying CPU and GPU parallel processing capabilities. 
Experimental results show that CSM-NN reduces the simulation time by up to $6\times$ compared to a state-of-the-art current source model based simulator running on a CPU. This speedup improves by up to $15\times$ when running on a GPU. CSM-NN also provides high accuracy levels, with less than $2\%$ error, compared to HSPICE. 
\end{abstract}

%% file: introduction.tex
\section{Introduction}
\label{sec:intro}




The down-scaling of transistor geometries has drastically increased the complexity of short channel effects and \textit{process-voltage-temperature} (PVT) variations.  
Consequently, \textit{application-specific integrated circuit} (ASIC) design flow techniques, such as \textit{multi-corner multi-mode} (MCMM) and \textit{parametric on-chip variation} (POCV) depend on increasingly more complex analysis, transformation, and verification iterations, to ensure the ASIC system functions correctly and meets design demands such as those related to performance, power and signal integrity. 
In these methods, the design is tested in different \textit{process-voltage-temperature} (PVT) corners and operating modes such as \textit{low-power} (LP),  \textit{high-performance} (HP), etc. 
Accurate simulation such as those for timing analysis during placement, clock network synthesis, and routing is crucial as it helps to lower the number of design iterations,  speed up convergence, and plays a major role in the turnaround time of complex designs such as \textit{system-on-chips} (SoCs)~\cite{Kahng-ICCD2018}. 

SPICE simulations are accurate but very slow for timing, power, thermal analysis, and optimization of modern ASIC designs with billions or trillions of transistors~\cite{thermal-pedram-2006, dynamic-power-benini-2000}. 
Therefore, higher levels of circuit abstraction using approximation has been used to speed up simulation steps. Abstraction models are generally based on \textit{look-up-tables} (LUTs), closed-form formulations, factors or their combinations. 
The traditional models, namely \textit{nonlinear delay model} (NLDM), \textit{nonlinear power model} (NLPM), \textit{effective current source model} (ECSM~\cite{Cadence-ECSM}), and \textit{composite current source model} (CCSM~\cite{Synopsys-CCSM}) utilize LUTs for storing delay, noise or power as nonlinear functions w.r.t. physical,  structural, and environmental parameters, and depend on voltage modeling more than current modeling. We refer to NLDM, ECSM, and CCSM models as \textit{voltage-LUT} (V-LUT) throughout this paper.
The V-LUT models are intuitively better choices when compared to simple closed-form formulation of nonlinear functions, however, tend to be increasingly inaccurate in capturing signal integrity and short channel effects with the down-scaling of technologies~\cite{goel-integrity-date2008}. 

Alternatively, current source models (CSMs) \cite{croix2003blade,Cadence-CSM,keller2004robust, goel2008statistical,amelifard2008current,knoth2012current,Shahin-TVLSI2011, fatemi2006statistical, fatemi2007current} 
use voltage-dependent current sources and possibly voltage-dependent capacitances to model logic cells. In addition to higher accuracy, another advantage of CSM over V-LUT models is the ability to simulate realistic waveforms for arbitrary input signals and provide the output waveforms. 

The number of CSM component values that should be stored in memory grows exponentially with the number of inputs and internal nodes in the logic cell. 
For example,  6-dimensional LUTs are required for modeling a 3-input NAND gate (NAND3). 
While V-LUT models are stored in smaller/faster memories such as L1-cache, relatively bigger tables in CSM-LUT can only fit into bigger/slower ones, like DRAM.  
Therefore a fundamental idea to shorten simulation time would be to replace some of the memorization with computation aiming for optimal space/time efficiency. 

In \cite{csm-cui-aspdac2014}, a \textit{Semi-Analytical CSM} (SA-CSM) was presented which uses small-size LUTs combined with nonlinear analytical equations to simultaneously achieve high modeling accuracy and space/time efficiency. However, developing analytical equations for complex circuits is a tedious process.

In this work, we propose CSM-NN, a circuit simulation framework that fully replaces LUTs with \textit{neural networks} (NNs). This eliminates the long memory access latency of LUTs, hence significantly shortens the simulation time, especially when CSM-NN computations can take advantage of parallelism offered by \textit{graphical processing units} (GPUs)~\cite{gpu-parallel-future}.

The major contributions of our work are as follows:
\begin{itemize}
    \item We developed a framework for simulating nonlinear behavior of complex integrated circuits using optimized NN structures as well as training and inference algorithms, according to the underlying CPU or GPU computational capabilities.
    \item Our framework is scalable and technology-independent, i.e., it can efficiently handle increasingly complex technologies with high PVT variations while maintaining the accuracy and improving the simulation latency. 
\end{itemize}
\vskip 5mm
The remainder of our paper is organized as follows. Section~\ref{sec:background} presents a short background on CSM and process variation issues. Sections~\ref{sec:methodology} and ~\ref{sec:experiments} elaborate our CSM-NN framework and experimental results, respectively. Section~\ref{sec:conclusion} concludes the paper. 
\vskip 5mm

%% file: csm.tex
\section{Background} \label{sec:background}
In this section, we briefly touch upon the basics of CSM and latency issues related to CSM-LUT memory access.



\begin{figure}[t]
\centering
\subfloat[CSM for single input (INV) logic gate.]{%
\includegraphics[width = 0.33\textwidth, keepaspectratio]{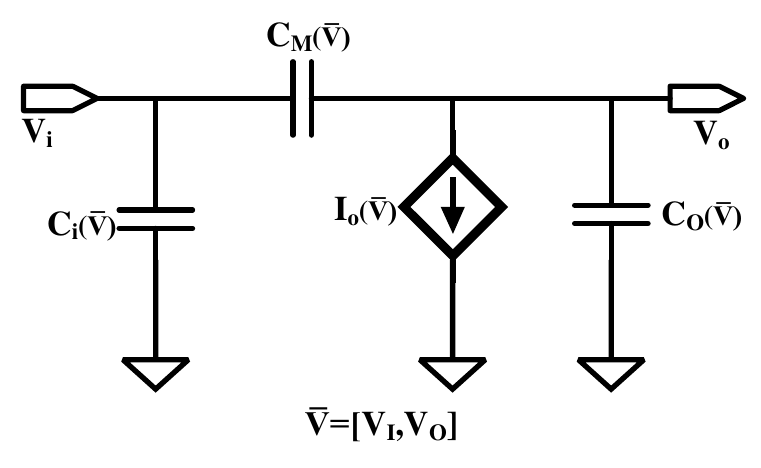}\label{a}}
\newline
\subfloat[CSM for two-input (NAND2) logic gate.]{%
\includegraphics[width = 0.40\textwidth, keepaspectratio]{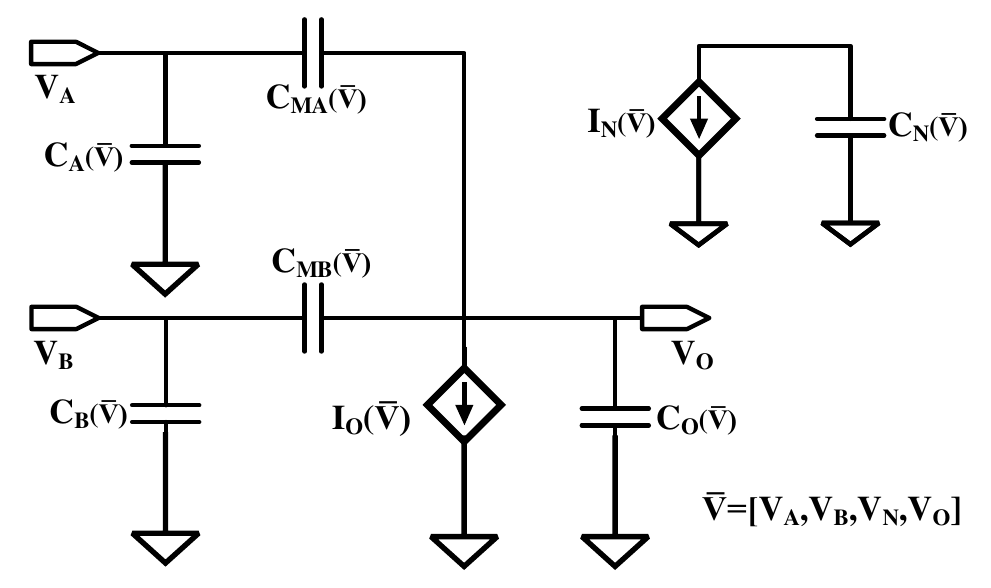}\label{b}}
\caption{CSM examples for one and two input logic cells~\cite{fatemi2006statistical, amelifard2008current}.}
\label{fig:csm-inv-nand}
\end{figure}

Each logic gate can be modeled using voltage-dependent current source as well as (miller and output) capacitance components~\cite{croix2003blade}. The values of these components can be characterized using HSPICE simulations. The CSM components of a logic cell can be stored in LUTs and utilized for noise, timing and power analysis of VLSI circuits~\cite{amelifard2008current, fatemi2007current, fatemi2006statistical, safar-PCA-date2010}.
Fig.~\ref{fig:csm-inv-nand} illustrates CSMs for single-input (INV) and multi-input (NAND2) logic cells.


Given a large number of simulation runs needed during the ASIC design and verification flow, and the corresponding long memory retrieval time, it is desirable to keep the number of dimensions and size of LUTs very small. Table~\ref{tab:csm-simple} lists the size of CSM LUTs for a simple library of basic gates. 

The size of CSM-LUTs for simple logic cells (c.f. Table~\ref{tab:csm-simple}) is an exponential function of logic cell complexity. As an example, NOR2 LUTs are 200 times larger than the one for INV, and XOR2 LUTs are 20,000 times larget than NOR2 ones. Note that in practical research or industrial standard cell libraries, there may be many logic cells of various sizes and complexities, some of which could be more complex than simple logic cells in Table~\ref{tab:csm-simple}. 

\vskip 5mm
\begin{table}
\caption {CSM for simple logic cells. Number of LUT dimensions ($\#Dim$), i.e. the count of inputs, outputs and internal voltage nodes; voltage-dependent capacitances ($C: C_M, C_o, C_i$) and current sources ($I:I_D, I_N$) required to model the cell; and the total size to be stored in memory ($LUT-Size$). All CSM-components are considered to be represented with 32bit (4Byte) floating points (FP). Characterization resolution is assumed to be 10 points per dimension.}
\centering
\label{tab:csm-simple}
\begin{tabular}{llll}
\textbf{Gate} & \textbf{\#Dim.} & \textbf{Variables} & \textbf{Table Size}\\ \hline
INV   & 2 & $3\times C, 1\times I$ &  $4 \times 10^2$ FPs = 1.6KB \\
NAND2 & 4 & $6\times C, 2\times I$ & $8 \times 10^4$ FPs = 320KB \\
NOR2  & 4 & $6\times C, 2\times I$ & $8 \times 10^4$ FPs = 320KB \\
AOI   & 6 & $9\times C, 3\times I$ & $12 \times 10^6$ FPs = 48MB \\ 
NAND3 & 6 & $9\times C, 3\times I$ & $12 \times 10^6$ FPs = 48MB \\ 
NOR3  & 6 & $9\times C, 3\times I$ & $12 \times 10^6$ FPs = 48MB \\
XOR2   & 8 & $12\times C, 4\times I$& $16 \times 10^8$ FPs = 6.4GB \\
\end{tabular}
\end{table}
Looking at the memory hierarchy details of Intel Broadwell micro-architecture~\cite{Broadwell} in Table~\ref{tab:cpu-spec} and comparing them with sizes in Table~\ref{tab:csm-simple}, confirms that CSM LUTs cannot fit in any of the caches and should be stored in the main memory (DRAM) and written into cache in parts. The latency of memory access in DRAM is about 2 orders of magnitude higher than that of L1 cache. 
This main difference shows the extent of longer  simulation latencies for CSM-LUT, compared to V-LUT. 

In the following two sections, we present how our CSM-NN eliminates the need for LUTs, and instead utilizes NNs to compute the CSM data. 


\vskip 5mm
\begin{table}[h]
\centering
\caption {Latency values for information retrieval from different hierarchy levels of memory and hardware specifications of Intel Xeon E5-2699 v4 server processor with Intel Broadwell micro-architecture. The computational capability of the processor is given in \textit{Giga floating point operations per seconds} (GFLOPs).} \label{tab:cpu-spec} 
\begin{tabular}{lll}
\hline \hline
\multicolumn{3}{c}{\textbf{Intel Broadwell micro-architecture}} \\ \hline 
\textbf{Memory} & \textbf{Size (KByte)}     & \textbf{Latency (Clock Cycle)} \\ 
L1 Data Cache   & 32     & 4-5 \\
L2 Cache        & 256    & 11-12  \\
L3 Cache        & 20,480  & 38 - 42 \\
DRAM            & -  & $\approx$ 250 \\ \hline \hline
\multicolumn{3}{c}{\textbf{Intel Xeon Processor E5-2699 v4}} \\ \hline 
\multicolumn{2}{l}{Cores} & 22 \\
\multicolumn{2}{l}{Base Frequency} & 2.2 GHz \\
\multicolumn{2}{l}{Single Precision}  & 774.4 GFLOPs \\
\multicolumn{2}{l}{Double Precision}  & 1548.8 GFLOPs \\ \hline
\\
\end{tabular}
\end{table}
\vskip 5mm


%% file: nn_modeling.tex

\section{CSM-NN Framework}\label{sec:methodology}
The description of our CSM-NN, including NN architecture and optimization algorithms for training is as follows.
\subsection{NN Architecture and Computation} \label{sec:neural-network}
To avoid the large LUTs with long query latencies in CSM-LUT, our CSM-NN, embeds parametric nonlinear models that can be trained on fully-connected NNs, to represent nonlinear functions. 

We believe CSM-NN can benefit from the following ML developments: (1) evolution of novel ML algorithms can be utilized towards improving the accuracy and efficiency of CSM-NN; and more importantly (2) exponential increase in computational capabilities, especially with recent advances in design of GPUs~\cite{Ng-ML-GPU}, significantly helps improving the performance of CSM-NN.

CSM-NN substitutes memory retrieval with computation, thus it is necessary to analyze and optimize the number of different structure and latency of operations required for CSM-NN in different hardware platforms.

There are two steps for CSM-NN: (1) simulation using a feed-forward pass that calculates the output of the model based on trained parameters and input values, and (2) back-propagation step, which modifies the parameters of the model based on the error, i.e. the difference between the expected values of the training data and the estimated output from the model.
Since the training process is done only once, computation during back-propagation is not a concern. 
Our objective is to improve the circuit simulation time. We therefore focus mainly on the inference process, i.e., we optimize the computation steps of the feed-forward pass. 

To choose the best NN architecture for our CSM-NN, we note that the number of hidden layers and the number of neurons in the hidden layer(s) determine the total number of parameters in the input-output function and the flexibility of the model.
Increasing the number of hidden layers beyond one (i.e., making the model \textit{deeper}) instead of increasing the number of neurons in a single layer (i.e., making the layer \textit{wider}) can also be considered. 
In \textit{deep neural networks} (DNNs), the sequence of nonlinear activation layers enables the input-output dependency to have a higher degree of nonlinearity with more flexibility. 
Although there are still unanswered questions on profound results of DNNs~\cite{deep-shallow-2017}, the belief is that multiple layers perform better at \textit{generalizing} as they learn the intermediate features between the raw input-data and the high-level output~\cite{deep-shallow-2017, DCNN-feature-2018}. 
As an example, thanks to the availability of data and computation resources in the past few years, the state-of-the-art solutions for challenging ML problems, such as image classification in the fields of computer vision, are made possible by creating models with over hundreds of layers~\cite{deeper-CVPR2015}~\cite{resnet-CVPR2016}. 
On the other hand, shallow networks do not generalize well but are very powerful in memorization~\cite{deep-shallow-2017}. 
In addition, training deeper models requires more data and time for training and also needs more computational resources for the feed-forward pass. 

In conclusion, despite the recent emergence of the DNN solutions and applications and potential improvement of accuracy of circuit simulation for complex timing, noise, and power analysis, we do not believe DNN is a feasible choice for the architecture of CSM-NN.

In the mathematical theory of \textit{artificial neural networks} (ANNs), the universal approximation theorem~\cite{universal-theorem} affirms that a single-hidden-layer NN can approximate continuous functions with a finite number of neurons, under assumptions over the nonlinear activation function and availability of sufficient data for training. 
Consequently, if a shallow wide network is trained with every possible input value, it could eventually memorize the corresponding output.
The following characteristics of our problem further suggest that  shallow wide networks with one hidden layer are more plausible solutions: 
\begin{itemize}
    \item There are no discontinuity in CSM component values. 
    \item While in practical applications the training data is limited or expensive to  generate, in CSM-NN it is straight forward to generate  training data with HSPICE simulations during the characterization process. 
    \item The number of inputs to the neural network is relatively small, even for complex logic cells, and when considering PVT parameters (Table~\ref{tab:csm-simple}). This implies that we are modeling a \textit{low dimensional} function. 
\end{itemize}

\begin{figure}[t]
\centering
\includegraphics[width = 0.46\textwidth, keepaspectratio] {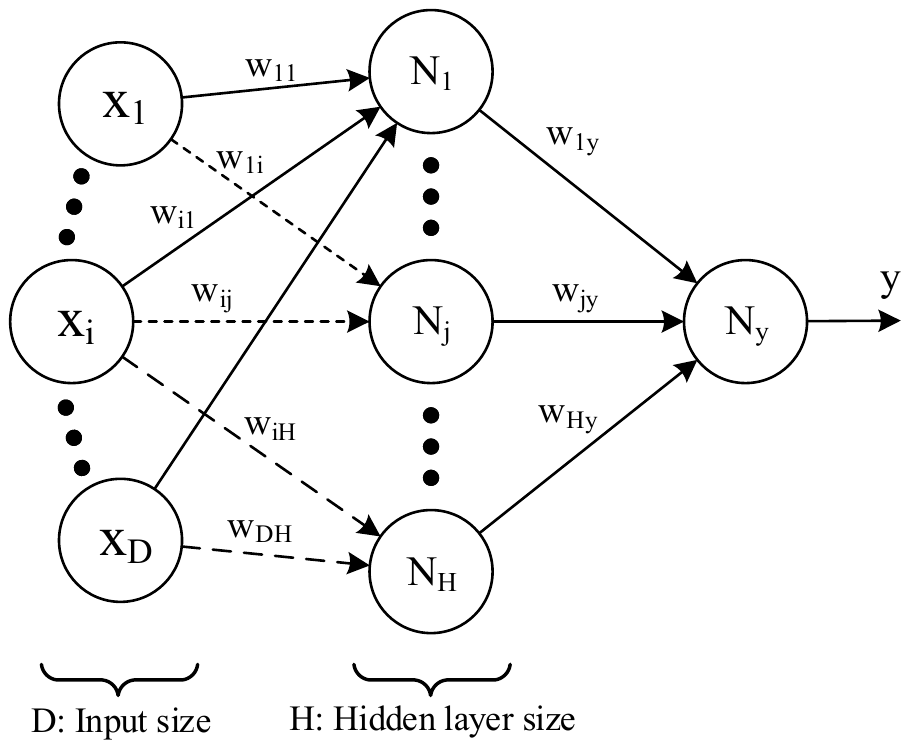}
\caption{One layer NN architecture used in CSM-NN. $x$, $y$, $w$, and $N$ are the inputs, output, weights, and the neurons respectively. The number of inputs (i.e., the dimension) and the width of the hidden layer are represented with $D$ and $H$, respectively.}
\label{fig:NN}
\end{figure}

Based on these features and considering the impact on inference step during circuit simulations, CSM-NN adopts a simple NN architecture with a single hidden layer to model the nonlinear behavior of CSM-NN components. The architecture and input-output function are shown in Fig.~\ref{fig:NN} and Eq.~\ref{eq:NN-formula}. 

\begin{equation}
\label{eq:NN-formula}
\begin{split}
\textbf{x} &= [x_0=1, x_1, x_2, ..., x_D] \\
g_i &= \sum_{d=0}^{D} w^{l=1}_{id}x_d = \textbf{w}_i^{l=1}\textbf{x} \\
a_i &= \sigma (g_i) = \frac{1}{1+e^{g_i}} \xrightarrow{} \textbf{a} = [1, a_1, ..., a_H] \\
y &= \sum_{h=0}^{H} w^{l=2}_{ih}a_h = \textbf{w}^{l=2}\textbf{a} \\
&= \sum_{h=0}^{H} \sigma (\sum_{d=0}^{D} w^{l=1}_{id}x_d)
\end{split}
\end{equation}

The number of MUL operations in feed-forward pass is equal to the number of model-parameters as calculated in Eq.~\ref{eq:num-weights}. It is very important to note that there are no dependencies among MUL steps in a specific layer, therefore they can be completely parallelized.

\begin{equation}\label{eq:num-weights}
    \text{\#MUL}=D\cdot H + H = (D+1)\cdot H 
\end{equation}

Considering notation used in Eq.~\ref{eq:NN-formula}, there are $H$ summations of $D$ values in the hidden layer. These $H$ summations also can be parallelized completely. To calculate the output, the summation of $H$ values is required. This summation can be efficiently parallelized by using tree-structures. The total number of ADD operations and the latency of tree-structure summations are calculated in Eq.~\ref{eq:num-adds} and Eq.~\ref{eq:latency-adds}. 

\begin{equation}\label{eq:num-adds}
    \text{\#ADD} = D \cdot H + H = (D+1)\cdot H 
\end{equation}

\begin{equation}\label{eq:latency-adds}
    \text{Latency} =  \ceil{log_2 D} + \ceil{log_2 H}
\end{equation}

\begin{figure}
\centering
\includegraphics[width = 0.48\textwidth, keepaspectratio] {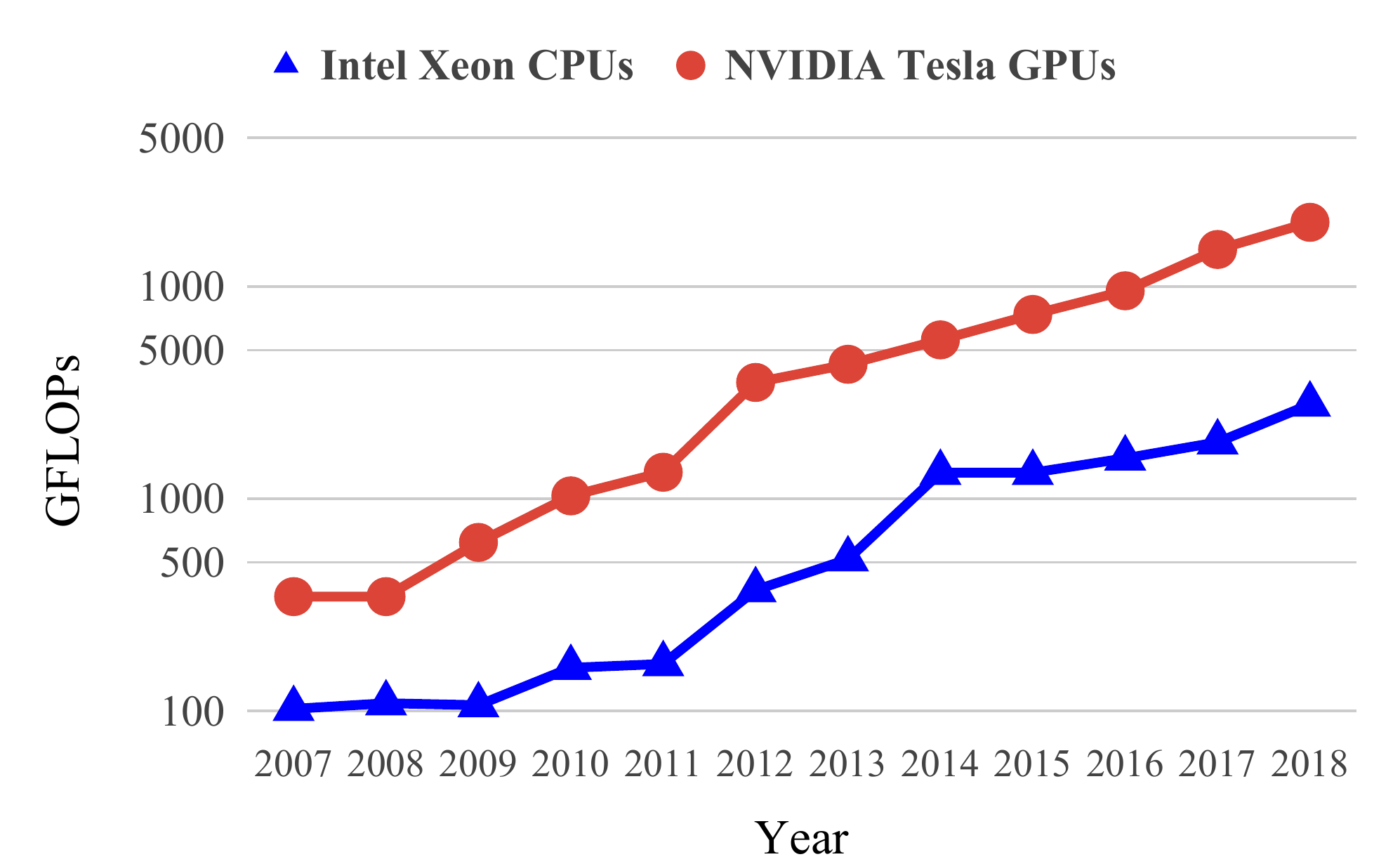}
\caption{Theoretical peak FLOPs with single precision.}
\label{fig:GPU-trend}
\end{figure}

CSM-NN accounts for the availability of resources when applying parallelization. NNs can be trained and utilized in two different hardware platforms, namely CPUs and GPUs. The evolution of GPUs and CPUs in case of \textit{number of floating-point operations per second} (FLOPS) are shown in Fig.~\ref{fig:GPU-trend}.

\subsubsection{CPU}
There are two phases of CSM-NN simulation computation when using CPUs: first, the weights of the NNs are loaded from the memory; and second, MUL and ADD operations are performed by \textit{arithmetic logic units} (ALUs). As later described in Section~\ref{sec:experiments}, the number of  CSM-NN parameters is sufficiently small. Therefore, they can fit into the cache (L1) of a CPU, and are accessible by the ALU in the order of a few CPU clock cycles. 

\subsubsection{GPU}
The computational capabilities of GPUs have increased dramatically in the past decade. This has made GPUs a good choice of hardware platform for NN computation~\cite{Ng-ML-GPU}.  

There are two levels of parallelized processing units in GPUs: 
several \textit{multiprocessors} (MPs), and several \textit{stream processors} (SPs, also referred as cores) that run the actual computation for each multiprocessor. 
Each core is equipped with ADD and MUL arithmetic units and designated register files. 
By implementing a trained NN (fixed parameters) on a GPU, the weights of each operation can be stored in register files, therefore, retrieval of information from memory is not required. 
We will show in Section~\ref{sec:experiments} that NNs of our CSM-NN framework can fit into a typical GPU. 
As an example, the hardware specifications of an NVIDIA GPU equipped with CUDA~\cite{CUDA-2008} cores is shown in Table~\ref{tab:gpu-spec}.

\begin{table}[h]
\caption {NVIDIA Tesla P100 GPU Specifications.} \label{tab:gpu-spec} 
\centering
\begin{tabular}{ll}
\hline
Streaming Processors (SM)           & 56 \\
32bit FP CUDA core (per SM / total)   & 64 / 3584 \\
64bit FP CUDA core (per SM / total)  & 32 / 1792 \\
Register file per SM & 256~KB \\
Shared memory per SM & 96~KB \\
Register file per CUDA core & 4~KB \\
Total L1 cache & 64~KB \\
Base clock frequency   & 1328~MHz \\
Single Precision GFLOPs & 9519 \\ \hline
\end{tabular}
\end{table}

It is worth noting that LUT-based models such as CSM-LUT and V-LUT models are only dependent on memory queries, thus using GPUs will not improve their simulation time. Therefore, considering relatively stronger parallelization capabilities of GPUs over CPUs, the speed advantage of CSM-NN over CSM-LUT and V-LUT improves, when running on GPUs. 


\subsection{Training Process}\label{sec:nn-training}
We have adopted L-BFGS as the optimization technique for training the NNs of our CSM-NN framework. The following provides our justification. 
There are several gradient descent based optimization algorithm candidates such as \textit{stochastic gradient descent} (SGD), Nesterov, Adagrad, and ADAM~\cite{deeplearning-Goodfellow-2016} to be considered for the training of neural regression models. SGD and inherited algorithms, such as ADAM, are by far the most popular algorithms to optimize NNs~\cite{SGD-survey}.
Their advantages to other techniques include parallelization, fast computation, and use of minibatch training techniques for better generalization specially in DNNs.
The functionality of these methods is conditioned to the appropriate tuning of hyper-parameters for training. 
On the other hand,  Quasi-Newton methods such as \textit{Broyden-Fletcher-Goldfarb-Shanno~} (BFGS), can be orders of magnitude faster than SGD. 
These methods are based on measuring the curvature of the objective function to select the length and direction of the steps. 
The main shortcoming of BFGS is that it requires high computation and memory resources when calculating the inverse of Hessian matrix for large datasets. 
\textit{Limited memory BFGS} (L-BFGS)~\cite{LBFGS} is an optimization algorithm in the family of quasi-Newton methods that approximates the BFGS algorithm using a limited amount of memory. 

The experimental results for low dimensional problems in~\cite{Ng-BFGS-2011} show that L-BFGS produces highly competitive or sometimes superior models compared to SGD methods. 
Another important advantage of L-BFGS is that it requires adjusting zero (and in advanced modified versions of L-BFGS, only a few) hyper-parameters. 
For example, differently from SGD, the learning rate (step-size) of L-BFGS  is tuned internally. We should also note that while several mini-batch versions of L-BFGS have been suggested very recently in the literature~\cite{lbfgs-batch-2018}, L-BFGS is generally considered as a batch algorithm and thus no batch-size adjustment is required. 
Considering these specifications, we chose L-BFGS as our optimization technique for training the NNs in the CSM-NN method. 


The common approach in supervised learning is to verify the generalization of the trained model by utilizing a validation (test) dataset which is completely separate from the training dataset. This process would prevent the possible over-fitting of the model. Therefore, we can randomly select samples from characterization data and test the accuracy of model.


It is very important to note that while accuracy of NNs in predicting CSM component values is important, the accuracy should ultimately be measured based on the quality of the output signal waveforms. 
Even the measurement of the propagation delay of the gate is not sufficient to confirm the accuracy of a CSM simulator. 
Therefore, similar to~\cite{amelifard2008current, ssta-csm-dac2016}, we used \textit{expected waveform similarity} ($\mathrm{E_{sim}}$) as a figure of merit for the measurement of the accuracy of our CSM simulations. 
In this work, $E_{sim}$ is defined as the mean of the absolute difference between precise HSPICE and CSM-NN simulations relative to the supply voltage value of the technology as shown in Eq.~\ref{eq:esim}. 

\begin{align}\label{eq:esim}
    \mathrm{E_{sim} = \frac{1}{T \times V_{DD}} \int_{0}^{T} |V_{ SPICE}-V_{ CSM-NN}|}
\end{align}

\subsection{CSM-NN Flow}
Technology information and standard cell libraries at the transistor level are provided by semiconductor manufacturers and design parties. 
Each of the cells in the standard library should be characterized separately for every PVT corner and mode settings. 
The number of different MCMM settings is technology and product design policy dependent. 
The characterization process is usually very time intensive, and can be done in different resolutions. 
While higher resolutions result in higher accuracies, they need a longer characterization times. 
It should be mentioned that more data needs a larger memory in CSM-LUT and possibly a longer training process in CSM-NN. 
Therefore, choosing an appropriate resolution is an important step in both CSM-LUT and CSM-NN flows. 
While our results in section~\ref{sec:experiments} are technology specific, they suggest a range of acceptable characterization resolutions. 
Up to this point of the flow, CSM-NN steps coincide with those of CSM-LUT. 

The next step is to train the NNs, one for every CSM component (e.g. $I_o$), 
of a logic cell ($INV$) and in a specific PVT corner (e.g. \textit{fast-fast and high temperature} (FFHT)). 
The inputs of the NNs are the voltages of terminal and internal nodes ($V_{I}, V_{O}$), and the target output is the value of the CSM components in these voltage points ($C_M(V_I, V_O)$). 

The training data collected through characterization should first be preprocessed and then used for training. 
As explained in section~\ref{sec:neural-network}, wider network can result in a more accurate model, but requires more computation.
Hence, we need to find an appropriate layer size. 
We choose the smallest number of neurons such that the network can pass a pre-defined accuracy threshold in terms of $E_{sim}$.

In the following section, we will show that this optimal set of NN parameters can fit into the cache (L1) of a typical CPU or the register files of a typical GPU. 
To simulate a circuit in a specific MCMM setup, the corresponding NN models of all logic cells in the standard library are loaded.


%% file: experiment2.tex
\section{Experiments and Simulation Results} \label{sec:experiments}

We implemented the simulator and the flow of our CSM-NN framework in Python. Our implementation 
is technology independent and can characterize, and create NN models with flexible configurable setup, for any given combinational circuit netlist. NN implementation and training are based on the Scikit-learn~\cite{scikit-learn} package. 

CPU and GPU devices introduced in Table~\ref{tab:cpu-spec} and Table~\ref{tab:gpu-spec} are used for comparison between two platforms, as both products are introduced in the same year (2016) and their current retail prices are in the same order (of about 5,000 USD). 
In this following, we discuss our experiments including challenges regarding our specific problem setup.

\subsection{Selected Technologies}
 In this work, for better evaluation of our CSM-NN including its technology independent characteristics, we performed our experiments on both MOSFET (16nm) and FinFET (20nm) device technologies from Predictive Technology Model~(PTM)~\cite{PTM} packages. Two device types namely \textit{low-standby power} (LP) and \textit{high performance} (HP) are used in our experiments~\cite{finfet-msa}.


As technology scales down, a growing number of physical and fitting parameters are needed to model PVT variations.
However as pointed out in~\cite{TED2016-UCB, Dadgour-ICCAD2008,Matsukawa-2009, IEDM2009-Stanford-SRAM}, only a few of them are dominant, i.e., developing simulation models that account for those dominant parameters while ignoring the rest, provides sufficiently high accuracy levels. 
Following these studies, we considered the most important process variation factors for defining a limited number of process corners. 
There is no process variation distribution information available for PTM technologies. 
Therefore, we followed the same approach used in~\cite{PV-MOSFET-2010} which studied the same devices as this work to define PVT corners. 

All distributions but temperature are considered normal (Gaussian) and reported as $N(\mu, \sigma)$ with ($\mu$) and ($\sigma$), representing mean and standard deviation respectively. The typical temperature value is considered as 27\textdegree{C} and the highest temperature ($+3\sigma$ variation) as 125\textdegree{C}. 
The information of the distribution for process variation parameters and the defined process corners for experiments are provided in Table~\ref{tab:ptm-pvt}.

\begin{table}[h]
\caption {Process (P), Voltage (V), and Temperature (T) variation distributions of technologies used in experiments. The values of process attributes are reported for NMOS/NFET devices. $t$ is representing \textit{oxide thickness} ($t_{ox}$) for MOSFET and \textit{Fin thickness} ($t_{fin}$) for FinFET. The distributions are all Normal($\mu, \sigma$) and represented as ($\mu, \sigma$).} \label{tab:ptm-pvt}
\centering
\begin{tabular}{l|l|l|l|l|l}
\multicolumn{1}{c}{} & \multicolumn{5}{c}{PVT Variation Distribution} \\ \hline \hline
\multicolumn{2}{l|}{Technology}  & $V_{dd} (v)$  &   $\Phi_M (ev)$ & $N_A (1e20)$ & $t (nm)$ \\  \hline
\multicolumn{2}{l|}{Fin-LP} & 0.9,0.05    & 4.6,0.23    & -      & 15,0.5 \\
\multicolumn{2}{l|}{Fin-HP} & 0.9,0.05    & 4.4,0.22    & -      & 15,0.5 \\ 
\multicolumn{2}{l|}{MOS-LP} & 0.9,0.05    & 4.6,0.23    & 2,0.1  & 1.2,0.04 \\ 
\multicolumn{2}{l|}{MOS-HP} & 0.7,0.035   & 4.4,0.23    & 2,0.1  & 0.95,0.03 \\ 
\multicolumn{6}{c}{} \\ 
\end{tabular}
\begin{tabular}{l|l|l|l|l|l}
\multicolumn{1}{c}{} & \multicolumn{5}{c}{PVT variation in pre-defined corners} \\ \hline \hline
Corner & $V_{dd}(v)$ & $T$(\textdegree{C}) & $\Phi_M (ev)$ & $N_A$ & $t(nm)$ \\ \hline
FF   & $+3\sigma$ & $0\sigma$ & $-3\sigma$ & $+3\sigma$ & $-3\sigma$ \\ 
SS   & $-3\sigma$ & $0\sigma$ & $+3\sigma$ & $-3\sigma$ & $+3\sigma$ \\ 
FFHT & $+3\sigma$ & $+3\sigma$ & $-3\sigma$ & $+3\sigma$ & $-3\sigma$ \\ 
SSHT & $-3\sigma$ & $+3\sigma$ & $+3\sigma$ & $-3\sigma$ & $+3\sigma$ \\ 
\end{tabular}
\end{table}


\subsection{Characterization}
The resolution of characterization process is a key factor in determining the accuracy of both CSM-LUT and CSM-NN simulations. 
While more data points increase the accuracy of both simulators, it comes with the cost of longer characterization process, larger tables in CSM-LUT, and longer training time in our CSM-NN. 
We therefore evaluate our CSM-NN framework under different resolutions. The results can also be later used towards suggesting a baseline for other technologies. 

It should be mentioned that CSM-components exhibit different sensitivity levels to different voltage-node variables. For example, $C_O$ seems to be more sensitive to $V_O$ than $V_I$ in INVX1, and it can be characterized with lower resolution for $V_I$ than $V_O$. Moreover, the sensitivity to resolution of characterization for one CSM-component should not be necessarily the same as the other component. For example, the range of change in $I_o$ value for a single $INVX1$ transition is from $\mu A$ to $nA$, while this is about only $50\%$ for $C_O$. 
The resolution can also vary based on the range of the voltage-node variable, e.g., higher resolutions for the noisy parts of the waveform (with higher frequencies of change) and lower resolutions for smooth parts of the waveforms.

However, for the sake of simplicity, we considered all voltage-node resolutions as similar.
As the units for different dimensions are different, we defined three different resolution setups as explained in Table~\ref{tab:char-res}. 
By comparing the preliminary results, \textit{normal} setup was found to be an appropriate resolution and the experiments were continued with this setup.

\begin{table}[!h]
\centering
\caption {Characterization resolution settings used in our experiments.} 
\label{tab:char-res} 
\begin{tabular}{l|c|c|c}
 & S: Soft & N: Normal & C: Coarse \\ \hline
Resolution (v)& 0.01 & 0.05  & 0.1 \\ 
\end{tabular}
\end{table}


\begin{table}
\centering
\caption {Choice of NN hidden layer size for single and two input logic cells.} 
\label{tab:cell-NN} 
\begin{tabular}{l|l|l|l|l|l}
& TT & FF & SS & FFHT & SSHT \\ \hline \hline
\multicolumn{6}{c}{MOSFET-HP 16nm} \\ \hline 
INV & 14 & 16 & 18 & 16 & 18 \\ 
NAND2 & 24 & 28 & 30 & 28 & 30 \\ \hline \hline
\multicolumn{6}{c}{MOSFET-LP 16nm} \\ \hline
INV & 20 & 20 & 24 & 22 & 26 \\ 
NAND2 & 28 & 32 & 30 & 32 & 32 \\ \hline \hline
\multicolumn{6}{c}{FinFET-HP 20nm} \\ \hline
INV & 20 & 20 & 20 & 26 & 24 \\ 
NAND2 & 34 & 30 & 34 & 36 & 36 \\ \hline \hline
\multicolumn{6}{c}{FinFET-LP 20nm} \\ \hline
INV & 20 & 20 & 20 & 26 & 20 \\ 
NAND2 & 30 & 36 & 36 & 40 & 38 \\ \hline \hline
\end{tabular}
\end{table}

\begin{table*}
\centering
\caption{CSM simulation results of a full adder circuit in both FinFET and MOSFET technologies. The simulation time improvements ($A$) is the ratio of the time required for CSM-LUT simulation over the one for CSM-NN. While CSM-LUT results would not improve if they were run on a GPU (instead of a CPU), the improvement results for the CPU and GPU implementation of CSM-NN are reported as $A_{CPU}$ and $A_{GPU}$ respectively. The hardware platform's specs are reported in Table~\ref{tab:cpu-spec} and~\ref{tab:gpu-spec}. $E_{sim}$ is the measure of accuracy introduced in Eq.~\ref{eq:esim}. }
\label{tab:res-FA}
\begin{minipage}{\textwidth}
\centering
\begin{tabular}{|c|c|c|c|c|c|c|c|c|c|c|c|c|c|}
\hline
- & \multicolumn{3}{c|}{\textbf{MOSFET-HP 16nm}} & \multicolumn{3}{c|}{\textbf{MOSFET-LP 16nm}} & \multicolumn{3}{c|}{\textbf{FinFET-HP 20nm}} & \multicolumn{3}{c|}{\textbf{FinFET-LP 20nm}} \\ \hline
Corner 	& $E_{sim.}$ & $A_{CPU}$ & $A_{GPU}$ & $E_{sim.}$ & $A_{CPU}$ & $A_{GPU}$ & $E_{sim.}$ & $A_{CPU}$ & $A_{GPU}$ & $E_{sim.}$ & $A_{CPU}$ & $A_{GPU}$ \\ \hline
Nominal & \textless 2\% & 9.3 & 16.8 & \textless 2\% & 9.3 & 16.8 & \textless 2\% & 6.9 & 15.1 & \textless 2\% & 8.6 & 16.8 \\ \hline
FF  	& \textless 2\% & 8.6 & 16.8 & \textless 1\% & 7.4 & 15.1 & \textless 2\% & 8.6 & 16.8 & \textless 2\% & 6.8 & 15.1 \\ \hline
SS      & \textless 2\% & 8.6 & 16.8 & \textless 2\% & 6.9 & 15.1 & \textless 1\% & 6.9 & 15.1 & \textless 1\% & 6.8 & 15.1 \\ \hline
FFHT 	& \textless 2\% & 8.6 & 16.8 & \textless 1\% & 7.4 & 15.1 & \textless 1\% & 6.8 & 15.1 & \textless 2\% & 6.6 & 15.1 \\ \hline
SSHT 	& \textless 2\% & 8.6 & 16.8 & \textless 1\% & 7.4 & 15.1 & \textless 1\% & 6.8 & 15.1 & \textless 1\% & 6.6 & 15.1 \\ \hline
\end{tabular}
\end{minipage}
\end{table*}


\subsection{Preprocessing and Loss Function Modification}
Mean Square Error (MSE, also referred as L2-norm error) is a commonly used regression loss function. It is simply the average of squared distances between our targets ($y_i$) and predicted values ($\hat{y_i}$). The loss function can also accommodate regularization term added to the loss function in order to prevent overfitting by shrinking the model parameters. The values of CSM-components vary in a large scale. For example, in INV, with  $V_{I}$, $V_{O}$ as variables, the DC current $I_{D}$ is in $micro-scale$ when both transistors are on, while in $nano-scale$ when one of them is off and the cell is leaking. 
The MSE-loss is a function of absolute error. Thus, by using this loss, the error in lower scale values will be less important compared to the higher scale values. To address this, we can log-transform the output, so the relative error will be used for loss calculation of the regression model as shown in Eq.\ref{eq:MSE}. An issue with such an adjustment is that some of the values are negative and this makes the log-transform more complicated. We simply resolved such issue with a simple shift of data toward positive values by subtracting all data points with their overall minimum $y_{min}$. 

\begin{equation}
\label{eq:MSE}
\begin{split}
MSE &= \frac{\sum_{i=0}^{N} (y_{i} - \hat{y_i} )^2}{N} \\
MSE_{log} &= \frac{\sum_{i=0}^{N} (log(y_{i}-y_{min}) - log(\hat{y_i} - y_{min}))^2} {N} \\
\end{split}
\end{equation}

The normalization of data in regression problems would help the solvers with faster convergence and better numerical stability. Hence, normalization of inputs and outputs is typically implemented inside the solver, such as that in the Scikit-learn package~\cite{scikit-learn}  used in our implementation.


\subsection{NN Size and Training for Logic Cells}
To select the size of the hidden layer for each model, we repeated the training process for various neuron numbers in the range of $10-50$.
Preliminary results in our experiments showed that the \textit{tanh} nonlinear function provides better outcomes compared to other functions such as \textit{sigmoid} and \textit{ReLU}. 
As mentioned in Section~\ref{sec:nn-training}, there is no hyper-parameter, e.g., no learning-rate or mini-batch size tuning is required in L-BFGS optimization. 

The total number of generated data points is 500 per gate. We trained the NN with 90\% of this data (5-fold cross-validation, 360 for training and 40 for validation) and then tested on the other 10\%. The split between training, validation, and test datasets was done in random. 

Next, we applied a few noisy input smaples to the cell and measured $E_{sim}$. 
The minimum size of the hidden layer that met $E_{sim} < 1 \%$ is chosen as the CSM-NN architecture for the logic cell in the specific MCMM setup. 
The complete results for the choice of architecture for INV and NAND2 are given in Table~\ref{tab:cell-NN} for different MCMM setups. 


\subsection{Circuit Simulation}
\begin{figure}
\centering
\includegraphics[width=0.45\textwidth, keepaspectratio] {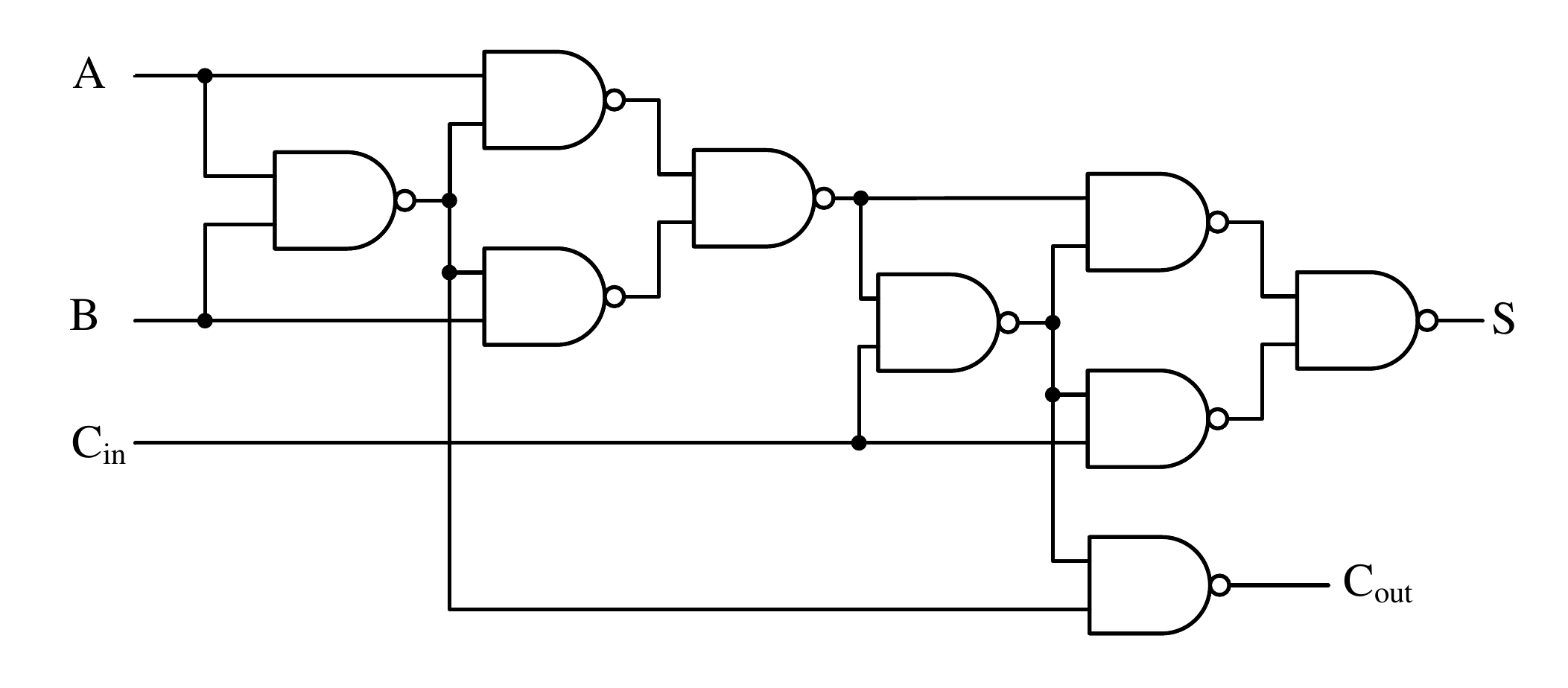}
\caption{Gate level schematic of the full adder circuit used in our experiments.}
\label{fig:full-adder}
\end{figure}

In this work we evaluated our CSM-NN framework by simulating a full-adder circuit (schematic shown in Fig.~\ref{fig:full-adder}). 

For the sake of a fair comparison, the HSPICE characterization setup is the same for both CSM-NN and CSM-LUT. 
We measured $E_{sim}$ by comparing output waveforms of HSPICE as the baseline with those of CSM-NN simulations. 
 The CPU and GPU devices used in our experiments are introduced in Table~\ref{tab:cpu-spec} and Table~\ref{tab:gpu-spec} respectively. 
CSM-LUT is considered to be computed on the CPU platform as it does not benefit from GPU parallelization. 
The required computation resources and latencies are calculated using equations in section~\ref{sec:neural-network}. 
The results confirm that CSM-NN output waveforms match those of HSPICE in regard to propagation delay with error values limited to 0.1\%. To better confirm the high accuracy of CSM-NN, we compared its waveform similarity to HSPICE, by measuring $E_{sim}$. As listed in Table~\ref{tab:res-FA}, $E_{sim}$ is limited to 2\%.

%% file: conclusion.tex
\vskip -2mm
\section{Conclusions and Future Work}\label{sec:conclusion}
CSM-NN, a scalable, technology-independent circuit simulation framework is proposed. 
CSM-NN is aimed to address the efficiency concerns of the existing tools that depend on data query from lookup tables stored in memory. 
Given the underlying CPU and GPU parallel processing capabilities, our framework replaces memorization by computation, utilizing a set of optimized NN structures, training and inference processing steps. 
The simulation latency of CSM-NN was evaluated in multiple MOSFET and FinFET technologies based on  predictive technology models in various PVT corners and modes. 
The results confirm that CSM-NN improves the simulation speed by up to $6\times$ using CPU platforms, compared to a CSM-LUT baseline. CSM-NN can further benefit from parallelization capabilities of GPUs, therefore the simulation speed is improved by up to $15 \times$ when run on a GPU. CSM-NN also provides high accuracy levels, maintaining the waveform similarity error within $2 \%$ compared to HSPICE. We believe the application of CSM-NN in future simulation tools such as those for sign-off and MCMM analysis and optimization of advanced VLSI circuits can significantly improve the simulation accuracy and speed.

As part of our future work, we plan to investigate CSM-NN on industrial circuits using accurate foundry technology information including PVT variations. 
We also plan to enhance our NNs to account for PVT corner parameters as inputs, to be able to train NNs once for all modes and corners and evaluate the cost vs speed and accuracy trade-off.

%% file: main.bbl
\begin{thebibliography}{10}
\providecommand{\url}[1]{#1}
\csname url@samestyle\endcsname
\providecommand{\newblock}{\relax}
\providecommand{\bibinfo}[2]{#2}
\providecommand{\BIBentrySTDinterwordspacing}{\spaceskip=0pt\relax}
\providecommand{\BIBentryALTinterwordstretchfactor}{4}
\providecommand{\BIBentryALTinterwordspacing}{\spaceskip=\fontdimen2\font plus
\BIBentryALTinterwordstretchfactor\fontdimen3\font minus
  \fontdimen4\font\relax}
\providecommand{\BIBforeignlanguage}[2]{{%
\expandafter\ifx\csname l@#1\endcsname\relax
\typeout{** WARNING: IEEEtran.bst: No hyphenation pattern has been}%
\typeout{** loaded for the language `#1'. Using the pattern for}%
\typeout{** the default language instead.}%
\else
\language=\csname l@#1\endcsname
\fi
#2}}
\providecommand{\BIBdecl}{\relax}
\BIBdecl

\bibitem{Kahng-ICCD2018}
A.~B. {Kahng}, U.~{Mallappa}, and L.~{Saul}, ``Using machine learning to
  predict path-based slack from graph-based timing analysis,'' in
  \emph{International Conference on Computer Design (ICCD)},  2018, pp.
  603--612.

\bibitem{thermal-pedram-2006}
M.~{Pedram} and S.~{Nazarian}, ``Thermal modeling, analysis, and management in
  {VLSI} circuits: Principles and methods,'' \emph{Proceedings of the IEEE},
  vol.~94, no.~8, pp. 1487--1501, Aug 2006.

\bibitem{dynamic-power-benini-2000}
L.~{Benini}, A.~{Bogliolo}, and G.~{De Micheli}, ``A survey of design
  techniques for system-level dynamic power management,'' \emph{IEEE
  Transactions on Very Large Scale Integration (VLSI) Systems}, vol.~8, no.~3,
  pp. 299--316, June 2000.

\bibitem{Cadence-ECSM}
\BIBentryALTinterwordspacing
{Cadence Inc., San Jose, California, U.S.} Cadence encounter library
  characterization datasheet. [Online]. Available:
  \url{https://www.cadence.com/content/cadence-www/en{\_}US/documents/Archive/Archive1/library{\_}characterizer{\_}ds.pdf}
\BIBentrySTDinterwordspacing

\bibitem{Synopsys-CCSM}
\BIBentryALTinterwordspacing
{Synopsys Inc., Mountain View, California, U.S.} {PrimeTime} datasheet.
  [Online]. Available: \url{https://www.synopsys.com/content/dam/synopsys/
  implementation{\&}signoff/datasheets/primetime-ds.pdf}
\BIBentrySTDinterwordspacing

\bibitem{goel-integrity-date2008}
S.~V. Amit~Goel, ``Current source based standard cell model for accurate signal
  integrity and timing analysis,'' \emph{Design, Automation and Test in Europe
  (DATE)}, pp. 574--579, 2008.

\bibitem{croix2003blade}
J.~F. {Croix} and D.~F. {Wong}, ``Blade and razor: cell and interconnect delay
  analysis using current-based models,'' in \emph{Design Automation Conference
  (DAC)}, 2003, pp. 386--389.

\bibitem{Cadence-CSM}
R.~Goyal and N.~Kumar, ``Current based delay models: A must for nanometer
  timing,'' \emph{Cadence Live Conference (CDNLive)}, 2005.

\bibitem{keller2004robust}
I.~{Keller}, {Ken Tseng}, and N.~{Verghese}, ``A robust cell-level crosstalk
  delay change analysis,'' in \emph{International Conference on Computer-Aided
  Design (ICCAD)},  2004, pp. 147--154.

\bibitem{goel2008statistical}
A.~Goel and S.~Vrudhula, ``Statistical waveform and current source based
  standard cell models for accurate timing analysis,'' in \emph{Design
  Automation Conference (DAC)}, 2008, pp. 227--230.

\bibitem{amelifard2008current}
B.~{Amelifard}, S.~{Hatami}, H.~{Fatemi}, and M.~{Pedram}, ``A current source
  model for {CMOS} logic cells considering multiple input switching and stack
  effect,'' in \emph{Design, Automation and Test in Europe (DATE)}, 2008, pp.
  568--573.

\bibitem{knoth2012current}
C.~{Knoth}, H.~{Jedda}, and U.~{Schlichtmann}, ``Current source modeling for
  power and timing analysis at different supply voltages,'' in \emph{Design,
  Automation Test in Europe (DATE)}, 2012, pp. 923--928.

\bibitem{Shahin-TVLSI2011}
S.~{Nazarian}, H.~{Fatemi}, and M.~{Pedram}, ``Accurate timing and noise
  analysis of combinational and sequential logic cells using current source
  modeling,'' \emph{IEEE Transactions on Very Large Scale Integration (VLSI)
  Systems}, vol.~19, no.~1, pp. 92--103, Jan 2011.

\bibitem{fatemi2006statistical}
H.~Fatemi, S.~Nazarian, and M.~Pedram, ``Statistical logic cell delay analysis
  using a current-based model,'' in \emph{Design Automation Conference (DAC)},
  2006, pp. 253--256.

\bibitem{fatemi2007current}
H.~{Fatemi}, S.~{Nazarian}, and M.~{Pedram}, ``A current-based method for short
  circuit power calculation under noisy input waveforms,'' in \emph{Asia and
  South Pacific Design Automation Conference (ASP-DAC)},  2007, pp. 774--779.

\bibitem{csm-cui-aspdac2014}
T.~Cui, Y.~Wang, X.~Lin, S.~Nazarian, and M.~Pedram, ``Semi-analytical current
  source modeling of {FinFET} devices operating in near/sub-threshold regime
  with independent gate control and considering process variation,'' in
  \emph{Asia and South Pacific Design Automation Conference (ASP-DAC)}, 2014,
  pp. 167--172.

\bibitem{gpu-parallel-future}
S.~W. {Keckler}, W.~J. {Dally}, B.~{Khailany}, M.~{Garland}, and D.~{Glasco},
  ``Gpus and the future of parallel computing,'' \emph{IEEE Micro}, vol.~31,
  no.~5, pp. 7--17, Sep. 2011.

\bibitem{safar-PCA-date2010}
S.~Hatami and M.~Pedram, ``Efficient representation, stratification, and
  compression of variational {CSM} library waveforms using robust principle
  component analysis,'' in \emph{Design, Automation and Test in Europe (DATE)},
  2010, pp. 1285--1290.

\bibitem{Broadwell}
\BIBentryALTinterwordspacing
Intel broadwell (2018) {CPU} micro-architecture specification. [Online].
  Available: \url{https://www.7-cpu.com/cpu/Broadwell.html}
\BIBentrySTDinterwordspacing

\bibitem{Ng-ML-GPU}
R.~Raina, A.~Madhavan, and A.~Y. Ng, ``Large-scale deep unsupervised learning
  using graphics processors,'' in \emph{International Conference on Machine
  Learning (ICML)}, 2009, pp. 873--880.

\bibitem{deep-shallow-2017}
H.~Mhaskar, Q.~Liao, and T.~Poggio, ``When and why are deep networks better
  than shallow ones?'' in \emph{AAAI Conference on Artificial Intelligence},
  2017, pp. 2343--2349.

\bibitem{DCNN-feature-2018}
T.~{Wiatowski} and H.~{Bölcskei}, ``A mathematical theory of deep
  convolutional neural networks for feature extraction,'' \emph{IEEE
  Transactions on Information Theory}, vol.~64, no.~3, pp. 1845--1866, March
  2018.

\bibitem{deeper-CVPR2015}
C.~{Szegedy}, {Wei Liu}, {Yangqing Jia}, P.~{Sermanet}, S.~{Reed},
  D.~{Anguelov}, D.~{Erhan}, V.~{Vanhoucke}, and A.~{Rabinovich}, ``Going
  deeper with convolutions,'' in \emph{Computer Vision and Pattern Recognition
  (CVPR)},  2015, pp. 1--9.

\bibitem{resnet-CVPR2016}
K.~He, X.~Zhang, S.~Ren, and J.~Sun, ``Deep residual learning for image
  recognition,'' in \emph{Computer Vision and Pattern Recognition (CVPR)},
  2016.

\bibitem{universal-theorem}
B.~C. Cs\'aji, ``Approximation with artificial neural networks,'' Master's
  thesis, Faculty of Sciences, Eötvös Loránd University, Hungary, 2001.

\bibitem{CUDA-2008}
\BIBentryALTinterwordspacing
J.~Nickolls, I.~Buck, M.~Garland, and K.~Skadron, ``Scalable parallel
  programming with cuda,'' \emph{Queue}, vol.~6, no.~2, pp. 40--53, Mar. 2008.
  [Online]. Available: \url{http://doi.acm.org/10.1145/1365490.1365500}
\BIBentrySTDinterwordspacing

\bibitem{deeplearning-Goodfellow-2016}
I.~Goodfellow, Y.~Bengio, and A.~Courville, \emph{Deep Learning}.\hskip 1em
  plus 0.5em minus 0.4em\relax MIT Press, 2016,
  \url{http://www.deeplearningbook.org}.

\bibitem{SGD-survey}
S.~Ruder, ``An overview of gradient descent optimization algorithms,''
  \emph{arXiv}, vol. abs/1609.04747, 2016.

\bibitem{LBFGS}
D.~C. Liu and J.~Nocedal, ``On the limited memory {BFGS} method for large scale
  optimization,'' \emph{Mathematical Programming}, vol.~45, no.~1, pp.
  503--528, Aug 1989.

\bibitem{Ng-BFGS-2011}
Q.~V. Le, J.~Ngiam, A.~Coates, A.~Lahiri, B.~Prochnow, and A.~Y. Ng, ``On
  optimization methods for deep learning,'' in \emph{International Conference
  on Machine Learning (ICML)}, 2011, pp. 265--272.

\bibitem{lbfgs-batch-2018}
R.~Bollapragada, D.~Mudigere, J.~Nocedal, H.-J.~M. Shi, and P.~T.~P. Tang, ``A
  progressive batching {L-BFGS} method for machine learning,'' in
  \emph{International Conference on Machine Learning (ICML)}, 2016.

\bibitem{ssta-csm-dac2016}
D.~Sinha, V.~Zolotov, S.~K. Raghunathan, M.~H. Wood, and K.~Kalafala,
  ``Practical statistical static timing analysis with current source models,''
  in \emph{Design Automation Conference (DAC)}, 2016, pp. 113:1--113:6.

\bibitem{scikit-learn}
F.~Pedregosa, G.~Varoquaux, A.~Gramfort, V.~Michel, B.~Thirion, O.~Grisel,
  M.~Blondel, P.~Prettenhofer, R.~Weiss, V.~Dubourg, J.~Vanderplas, A.~Passos,
  D.~Cournapeau, M.~Brucher, M.~Perrot, and E.~Duchesnay, ``Scikit-learn:
  Machine learning in python,'' \emph{Journal of Machine Learning Research},
  vol.~12, pp. 2825--2830, Nov. 2011.

\bibitem{PTM}
``{Predictive Technology Model} from arizona state university,''
  \url{http://ptm.asu.edu/}, accessed: 2019-05-20.

\bibitem{finfet-msa}
M.~S. {Abrishami}, A.~{Shafaei}, Y.~{Wang}, and M.~{Pedram}, ``Optimal choice
  of {FinFET} devices for energy minimization in deeply-scaled technologies,''
  in \emph{International Symposium on Quality Electronic Design (ISQED)},
  2015, pp. 234--238.

\bibitem{TED2016-UCB}
X.~{Zhang}, D.~{Connelly}, P.~{Zheng}, H.~{Takeuchi}, M.~{Hytha}, R.~J.
  {Mears}, and T.~K. {Liu}, ``Analysis of 7/8-nm {Bulk-Si} {FinFET}
  technologies for {6T-SRAM} scaling,'' \emph{IEEE Transactions on Electron
  Devices}, vol.~63, no.~4, pp. 1502--1507, April 2016.

\bibitem{Dadgour-ICCAD2008}
H.~{Dadgour}, {Vivek De}, and K.~{Banerjee}, ``Statistical modeling of
  metal-gate work-function variability in emerging device technologies and
  implications for circuit design,'' in \emph{International Conference on
  Computer-Aided Design (ICCAD)},  2008, pp. 270--277.

\bibitem{Matsukawa-2009}
T.~{Matsukawa}, S.~{O'uchi}, K.~{Endo}, Y.~{Ishikawa}, H.~{Yamauchi}, Y.~X.
  {Liu}, J.~{Tsukada}, K.~{Sakamoto}, and M.~{Masahara}, ``Comprehensive
  analysis of variability sources of {FinFET} characteristics,'' in
  \emph{Symposium on VLSI Technology},  2009, pp. 118--119.

\bibitem{IEDM2009-Stanford-SRAM}
{Xiao Zhang}, {Jing Li}, M.~{Grubbs}, M.~{Deal}, B.~{Magyari-Köpe}, B.~M.
  {Clemens}, and Y.~{Nishi}, ``Physical model of the impact of metal grain work
  function variability on emerging dual metal gate {MOSFET}s and its
  implication for sram reliability,'' in \emph{International Electron Devices
  Meeting (IEDM)},  2009, pp. 1--4.

\bibitem{PV-MOSFET-2010}
Y.~{Li}, C.~{Hwang}, T.~{Li}, and M.~{Han}, ``Process-variation effect,
  metal-gate work-function fluctuation, and random-dopant fluctuation in
  emerging {CMOS} technologies,'' \emph{IEEE Transactions on Electron Devices},
  vol.~57, no.~2, pp. 437--447, Feb 2010.

\end{thebibliography}
